# Incorporating planning intelligence into deep learning: A planning support tool for street network design

Zhou Fang, Ying Jin and Tianren Yang*

*Martin Centre for Architectural and Urban Studies, University of Cambridge, Cambridge CB2 1PX, UK*

*corresponding author at: Martin Centre for Architectural and Urban Studies, Department of Architecture, University of Cambridge, Cambridge, CB2 1PX, UK; ty290@cam.ac.uk

# Incorporating planning intelligence into deep learning: A planning support tool for street network design

Deep learning applications in shaping ad hoc planning proposals are limited by the difficulty in integrating professional knowledge about cities with artificial intelligence. We propose a novel, complementary use of deep neural networks and planning guidance to automate street network generation that can be context-aware, example-based and user-guided. The model tests suggest that the incorporation of planning knowledge (e.g., road junctions and neighborhood types) in the model training leads to a more realistic prediction of street configurations. Furthermore, the new tool provides both professional and lay users an opportunity to systematically and intuitively explore benchmark proposals for comparisons and further evaluations.

## Introduction

In the era of information and communication technologies, deep learning tools are increasingly being used to inform urban planning and place-making. Building upon emerging, open-source spatial data and planning needs for urban smartness and sustainability, deep-learning-oriented tools have been widely applied from regional-scale hazard detection (Rahmati et al., 2019) to city-level land-use classification (Hagenauer and Helbich, 2012) and street-level landscape evaluation (Middel et al., 2019).

Despite the rapid theoretical development and practical prevalence of deep learning models, their applications in the urban planning field are more constrained to providing perceptions of cities (e.g., site analyses) rather than shaping solutions. One of the main barriers lies in the over-reliance of deep learning models on algorithms in a "black box," with limited consideration of professional knowledge in guiding urban development (Castelvecchi, 2016; König et al., 2017).

A central question is how to integrate domain knowledge (e.g., urban planning and design) with deep learning models for reciprocity in support of planning-related tasks. This paper aims to fill the gap by proposing a novel method that combines deep learning algorithms and subjective planning knowledge for generating street network proposals. A conventional deep neural network was first trained based on the fundamental context of the surrounding street networks, topography, and slope aspect. The addition of planning-related guidance (e.g., the location of major transport junctions and street network pattern type annotations) in model training was examined to explore whether the improved model can better detect, recognize and capture the local street network features and underlying relationships. Based on a well-trained deep learning model that considers both the site context and planning guidance, changes in design requirements were made to generate alternative street networks that were shown to fit specific planning needs.

The model performance and proposal generations were explored in a case study area of Barcelona (with iconic, grid-like street networks in the center and irregular road alignments farther afield) and Prague (with an even richer mixture of street types, including modernist and socialist legacies as well as substantial medieval-built forms). Results showed that the incorporation of planning intelligence into deep learning can provide more realistic planning options and enrich alternatives in the early stage of urban planning and design.

The complementary use of domain knowledge and deep learning contributes to the literature and practice of planning support systems (PSS). On the one hand, the proposed framework provides situation-cogent evidence in support of local planning and decision-making that well maintains the urban historical and cultural context as well as local development requirements. The awareness of both the development context and professional knowledge meets the recent calls for sentience in theoretical advancements of PSS (see, e.g., Deal et al., 2017). On the other hand, this study makes it possible for users (e.g., community groups and decision-makers) with less domain-specific expert knowledge to easily and instantly generate a set of benchmark proposals for comparisons and further evaluations. The automatic generation of related proposals offers a substantial step forward in support of planning decisions for local communities and professional practitioners. These intuitive and content-rich alternatives are essential, particularly when a rapid design response is required (see, e.g., König et al., 2017).

**Related Works**

PSSs integrate theories of planning, computing, and spatial representation within a single framework to support strategic planning activities (Harris and Batty, 1993). The central objective of a planning support tool is to lead planners and decision-makers toward useful alternative proposals for comparative evaluations and, therefore, suggest better solutions (Yamu et al., 2017). For instance, to ensure cities operate smoothly, decisions on extension, demolition and other alterations of the urban environment need to be made rapidly, and the time windows for planning and designing streets and roads are narrow (Jin et al., 2013). The availability of PSS will effectively elevate the productivity of urban designers and engineers (Ross and Rodriguez, 1963; Pan and Deal, 2019), particularly in countries experiencing unprecedented waves of urbanization.

    Proposal generation and its spatial representation play an integral part in the development of PSS. A geographic information system (GIS) has long been used as a general-purpose tool to store, manage, and display spatial data for specific planning support tools (Harris and Batty, 1993; Geertman et al., 2013). Based on a GIS, an aggregate summary of statistical variables and simulation results from a PSS can be

displayed in a choropleth map, cartogram, and even a three-dimensional virtual environment (Batty, 1997).

With the emergence of deep learning techniques, procedural and example-based modeling have been increasingly applied to support automatic content generation and visualization for planning decisions (Hartmann et al., 2017). Procedural modeling relies on manually designated rule sets to produce proposals. Parish and Müller (2001) made one of the first attempts to generate three-dimensional city models for visualization using procedural approaches, where a Lindenmayer system was used to grow road networks and buildings conditioned on global goals and local constraints. Given an initial and a final road point, Galin et al. (2010) developed a cost minimization function to automate path creation, considering the slope of the terrain and natural obstacles. The function was then extended to generate hierarchical road networks between towns at a regional level (Galin et al., 2011). Similar procedural principles can also be applied to allocate land use, subdivide blocks and generate buildings (see, e.g., Chen et al., 2008; Lyu et al., 2015).

In comparison, example-based approaches learn from real-world cases in a pre-processing step to extract features and adopt them as templates. Hartmann et al. (2017) developed an automatic road generation tool, StreetGAN, using a generative adversarial network (GAN) to synthesize street networks in a fix-sized region that can maintain the consistency of urban layouts learned from the training data set. Similarly, Kempinska and Murcio (2019) trained Variational Autoencoders (VAEs) using images of street networks derived from OpenStreetMap to capture urban configurations using low-dimensional vectors and generating new street networks by controlling the encoded vectors.

Despite increasing efforts to introduce deep learning techniques for addressing planning concerns, current deep learning tools fall short of the requirements of a PSS. Although procedural modeling relies on domain knowledge to produce design proposals, the rules set for the generation are mostly general and abstract, without details to support context-specific awareness in place-making (Vanegas et al., 2009; Jiao and Dillivan, 2013). In comparison, the example-based approaches synthesize results as a pure common pattern from the training data input, which is insufficient to provide alternatives under different user-defined scenarios in planning practice.

This research aims to take advantage of both example-based approaches (to preserve structures in real-world examples) and procedural-based methods (to adapt generated content under user-defined rules) in proposal generation and spatial representation for street network design. More specifically, our research upgraded an existing example-based street network generation approach from a data synthesizing system (generate street network to imitate provided training data) to a context-aware generation system. This context-aware generation system was inspired by an active computer vision task, namely image completion (also known as image inpainting), to fill in the missing parts of an image in a content-aware way (Bertalmio et al., 2000). Our system incorporates state-of-the-art neural network structures for the image completion task (Pathak et al., 2016; Iizuka et al., 2017; Yu et al., 2018). This upgrade allows our model to recognize various types of street patterns within the context region. The model automatically generates appropriate street networks within the predefined region and reasonably adapts the generated street network to the surrounding street network. To incorporate the procedural-based theory into our model, a conditional generative adversarial network (Mirza and Osindero, 2014; Isola et al., 2017) was adapted to allow users generating street networks to insert major transport junctions and provide street network pattern type annotations.

**Study Area, Data and Method**

*Study Area*

To illustrate the working mechanism of our proposed system, two cities, Prague, the Czech Republic, and Barcelona, Spain, were selected to establish a street network database to train and test the performance of our tool. These two cities were selected not only for their rich mixture of different street network pattern types but also because of their similar and representative city expansion processes in Europe. The cities began from a medieval core, grew in carefully designed orthogonal grid regions, and expanded to surroundings with irregular streets and roads at the urban fringe. The expansion patterns resulted from the cities' medieval histories, hilly terrain, and requirements of connecting to the regional expressway.

*Data Preparation*

The street network database proposed was prepared with a series of pattern-related attributes such as road hierarchies, topography, junctions, and neighborhood types. We used multiple overlapping layers, or "channels," of an image to represent the comprehensive attributes. These channels were divided into two specific types: fundamental context and planning guidance. Fundamental context is a geographic representation of roads, including (1) a hierarchical street network, (2) elevation, and (3) the aspect of slope across the case study cities. OpenStreetMap, an open-source platform, was used as a vector representation of street networks with defined road hierarchies (e.g., motorway, primary, secondary, tertiary, and residential). Publicly available digital elevation data was sourced from the shuttle radar topography mission,[1] upon which the aspect of the slope was calculated.

Planning guidance reflects the planning intelligence that can potentially be incorporated into the design of street networks, including (1) the location of road junctions and (2) the archetypal street patterns. As shown in Figure 1, six archetypal street patterns are predefined and manually labeled across the study area: linear development, gated compound, medieval, irregular grid, and orthogonal grid.

[insert Figure 1 here]

To avoid model overfitting during the training stage and guarantee that our trained model has never been exposed to the data used for testing, we use Prague data to train the model and Barcelona data to test. More specifically, we randomly cropped 46,856 unrepeated multi-channel patches in a size of 256 × 256 pixels (2 × 2 m per pixel and 512 × 512 m per patch) within a prepared 6 × 10 km multi-channel city map of Prague and used these patches for model training. Note that the cropped training patches also include a random-generated noise channel and a mask channel. These two channels were used to define the context region and the street infilling region. The test data sets (26,872 unrepeated patches) were prepared using the same methodology within a 6 × 10 km region in Barcelona.

*Model Structure and Training*

We designed a context-aware, conditional generative model for street network generation to support planning decision-making. Under general principles of GANs, a generator (G) to produce street networks within a defined region (conditioned on the fundamental context and planning guidance) and a discriminator (D) to evaluate the generalized patterns are iteratively trained until the outputs and the ground truth are hard to distinguish. Two losses, namely generation loss[2] and adversarial loss[3], are jointly minimized to train the proposed deep neural network structure by iteratively updating the parameters within the G and D. Using a mixture of the two loss functions allows for stable training of a high-performance network model (Pathak et al., 2016). An adaptive learning rate optimization algorithm is adopted for optimization. The trained G is then used for performance evaluation and further applications.

---

[1] Provided by a joint endeavor of the National Aeronautics and Space Administration, the National Geospatial-Intelligence Agency, and the German and Italian Space Agencies.

The input of the G is a series of multi-channel patches composed of fundamental context and planning guidance of street networks, with a random-generated mask channel and masked noise channel defining the generation region. The output is a single-channel raster image that infills street networks within the defined regions. This raster image is then fed into an attached D with planning guidance (seen as conditions) to discern whether a street network is real (ground truth). The output of the D is the possibility of the input street network to be real.

[insert Figure 2 here]

---

[2] To quantify the total pixel-to-pixel difference between the generated street network and the real-world network within the generation region.

[3] Adversarial loss is considered and included in the loss functions of most image completion algorithms. By taking adversarial loss into account, the standard pixel-to-pixel loss minimization process can be turned into a min-max optimization problem in which the discriminator is jointly updated with the generation network at each iteration. This is crucial for our approach, due to the existence of multiple possible solutions.

**Experiments and Results**

*Experiment Design*

To evaluate the proposed planning support model for street network generation and to further illustrate its working mechanism, three progressive models were designed, as summarized in Table 1.

[insert Table 1 here]

Model 1 focuses merely on the fundamental context, leveraging the model power alone to generate street networks. Model 2 and Model 3 build on Model 1 but add planning guidance to its training and application.

*Settings for Model Training and Testing*

Our models for all designed experiments were trained with Python v3.7.1, PyTorch v1.1.0, CUDA v10.0 and CUDNN v7.3.1 on a single NVIDIA Tesla P100 GPU for 100 epochs on each experiment. Each training epoch took 5,857 iterations with a batch size of 8, and each iteration consumed 1.36 s to compute. The entire training procedure for each experiment took roughly 221.3 h (9.2 d).

For testing, the model ran at 0.21 s per data pair on the same GPU used for training and 3.16 s on Intel Xeon CPU @ 4.00GHz for data with a resolution of 256 × 256 pixels on average, regardless of the size of the generation region.

*Testing Results*

In Figures 3–5, a series of results of Model 1 to Model 3 with corresponding input data pairs and ground truth outputs are listed for visual comparison. Due to space limitations, we concatenate the surrounding street networks channel, junction nodes channel, and street pattern-type channel into one RGB image and list this as the input of the test. The elevation channel and aspect of the slope channel are hidden. All reported results are direct outputs from the trained model without any post-processing. More results are shown in the supplementary material.

Without any planning guidance, the generation of street networks by Model 1 seems surprisingly convincing. The trained model is capable of considering the surrounding context and, therefore, generating street networks consistent with the vicinity through appropriate connections and coherent hierarchies.

[insert Figure 3 here]

From the case E1 in Figure 3, we can notice that the trained Model 1 performs well when generating the road network within central Eixample, Barcelona. This is because the features of the road network there are simple and have a strong coherence. From the cases E2 to E4, we can see that although all the results are different from the ground truths, the results themselves are natural and realistic (difficult for people to discern whether the road network is an actual one or a generated one). Considering that the trained module generates street networks purely by relying on the surrounding street network and basic geographic information, all these results can be considered acceptable and welcome at this stage. Once guidance from the ground truths is incorporated, the street networks generated become more like the ground truths by fulfilling the constraints.

[insert Figure 4 here]

Figure 4 shows several representative results of Model 2, where we introduce the randomly maintained street junction nodes as prerequisites to guide the generation process. In all results listed, the module successfully connected all the provided junction nodes and adapted the generated network to the surrounding network. As shown in E1 to E4 in Figure 4, when generating a grid-like street network (orthogonal grid and irregular grid), the junction nodes can be used to predefine the orientation and size of the street block.

[insert Figure 5 here]

In Figure 5, given additional guidance from pattern-type annotations, our full model is capable of reproducing all five types of street patterns in a specified region with reasonable connectivity to the surrounding street network. The module also can generate mixed types of patterns within the generation region (E6).

To quantitatively compare the performance of our models on the entire testing data set, we report results of the evaluation in terms of mean $l_1$ error and mean $l_2$ error. As shown in Table 2, with additional guidance information gradually fed into models 1 to 3, they performed ever better regarding mean $l_1$ and $l_2$ errors. The additional guidance, junction nodes, and pattern-type annotations act as constraints to the model. These limitations screen out some potential ways to complete the street network within the examined region, guiding the generated street network toward the ground truth.

[insert Table 2 here]

Moreover, we found that the coverage of the provided context determines the performance of all three models (Model 1 to Model 3). The results of the $l_2$ errors of all samples in the testing set from Model 1 to Model 3 are plotted against their context coverage in Figure 6. Highly uncertain performance should be expected when the coverage of context is less than 40%. With a higher coverage of the context region provided, all models obtained better performance in terms of $l_2$ errors.

[insert Figure 6 here]

More samples with a 50% to 60% context region coverage achieved lower $l_2$ errors on Model 2 and Model 3 compared to that of Model 1. This finding demonstrates the necessity of providing planning guidance when generating a street network for a limited context street network. The positive effect of providing junction nodes and pattern-type annotations is clearly shown when we comparing the three 2-dimensional histograms side-by-side. As shown in Figure 6, the distribution of data samples with 50% to 60% context coverage on Model 2 and Model 3 becomes more concentrated at a 3% to 4% $l_2$ error level compared to that of Model 1 (the light green section is squeezed toward lower $l_2$ error). Thus, increasing context coverage with introduced junction nodes and pattern-type annotations results in more samples in the testing data achieving better results.

*Incorporating Planning Intelligence*

Based on the trained Model 2 and Model 3, we studied the cases where a series of planning guidance, junction nodes, and street pattern type annotations were introduced to produce a series of alternative results within seconds. We then attempted to extend the proposed planning support model into a human-computer interactive design tool.

By introducing junction nodes within the generation region at desired locations (Model 2), users can (1) anchor the generated street network at certain pre-defined locations, (2) use a series of junction nodes to guarantee that the generated street network can successfully bypass buildings or trees which have been agreed upon to preserve, and (3) use a different density of junction nodes to control the size of divided land parcels and the density of the generated street network.

The series of alternative street networks generated by inserting and moving street junction nodes within the generation region are shown in Figure 7. Model 2 successfully (1) predicted the existence of additional junctions apart from the ones provided, and (2) automatically discarded some inappropriate user-inserted junction nodes based on the surrounding context and knowledge gained from samples in the training data set. The generated street networks are reasonable and faithfully follow the planning guidance provided.

[insert Figure 7 here]

Figure 8 shows the generative capability of Model 3. By assigning different pattern type annotations and correspondence junction nodes, the model can generate different street network patterns within the predefined generation region with identical fundamental context. This makes it possible for users to introduce new types of street network patterns to a region and produce a series of novel street network designs for further assessments. The training database may be expanded by including more street network patches extracted from other major cities around the world. With more iconic street pattern types, our model can be trained to generate richer and more diverse types of street networks in developing regions according to the knowledge obtained from the training data and the series of planning guidance provided. An expansion of our database will easily and swiftly equip the proposed planning support tool with more comprehensive information and knowledge in street network planning.

[insert Figure 8 here]

Moreover, all user-guided generation processes mentioned above take less than half a second on the same GPU used for training the model. This swift response guarantees a satisfactory human-computer interaction experience and makes it possible for users with less domain-specific expert knowledge to generate hundreds of desired street network designs within five minutes. The ability to easily add data and planning knowledge as well as the swift response time, suggests our model is feasible of executing plans of higher quality and savings in both time and resources. The model's intuitive and interactive application experience also demonstrated its significant potential to providing easy-to-generate and user-friendly education and public engagement experience.

**Conclusions and Future Work**

In large and complex cities, coordination between different interventions on buildings, streets, roads, public transport and utility infrastructure is difficult. Extensive future urban growth is expected to make the already complex urban structures and fabric even more challenging to understand, plan, and design. To achieve an ad hoc street network infill or expansion, some planners and decision-makers prefer to learn from the urban historical context to retain the characteristics of place-making, while the others tend to propose functional street patterns to improve transport efficiency. However, current planning and design decision are made simply, with few alternative options for systematic comparison. This situation is far from adequate since the geometric and spatial configurations of streets and roads must reflect intricate, multi-dimensional, functional, aesthetic, social, and environmental requirements.

Against this backdrop, this research has made one of the first attempts to incorporate knowledge about cities into a deep learning framework for automatic street network generation, which hinges on both the fundamental context (i.e., surrounding street networks, topography and slope aspect) and planning guidance (i.e., transport junctions and block types). Three encoder-decoder networks (coupled with an adversarial network) were progressively tested, with different levels of incorporation of the above-mentioned components. The results suggest that: (1) all models can detect and automatically fit into the different types and hierarchies of road patterns in the local context, and the introduction of planning guidance leads to better generative performance; (2) the generative capability is positively associated with the size of the coverage area representing the surrounding ground truth; and (3) users can propose and explore various proposals by defining the locations of major transport junctions and types of street patterns, where inappropriate inputs (i.e., road junctions) can be automatically noted by the model.

The proposed models can provide planners and decision-makers an opportunity to systematically explore potential planning options, through predicting street networks' future expansion or infill based on the areas of the immediate vicinity (context-aware), existing precedents (example-based) and professional intelligence (user-defined or procedural-based). More specifically, various planning scenarios can be exploratively incorporated with the deep learning techniques to automatically and interactively infer urban layouts, which can suggest new, useful alternatives for proposal comparisons. This interactive automation process can be linked to further evaluation tools to explore, for instance, how different road patterns will influence traffic congestion (Lindorfer et al.), commuting behavior (Hu et al., 2020), rental market (Yang et al., 2019) and social equity (Cao and Hickman, 2019).

Although this research is promising, explorative pathways for future work remain. First, the data sets for training can be expanded to incorporate all major cities globally to better understand the diversity in street network patterns. The open-sourced map (i.e., OpenStreetMap) provides not only an opportunity for massive data training resources but also serves as a fair global platform for performance comparisons across different planning support tools. More best practices in urban configurations can also be extracted and circulated. Second, more planning-related elements can be incorporated into the proposed framework, including building layout, land use, urban amenities, and population and employment density. As these urban elements co-determine urban morphology (Lim et al., 2015), their inclusion will lead to a more integrated planning support tool, where users can share insights from the underlying relationships among different components in street configurations.

Table 1. Alternative models for training and testing.

|  | Channel | Model 1 | Model 2 | Model 3 |
|---|---|---|---|---|
| Input: fundamental context | Surrounding street networks | √ | √ | √ |
|  | Elevation | √ | √ | √ |
|  | Aspect of slope | √ | √ | √ |
| Input: planning guidance | Junction nodes | – | √* | √* |
|  | Street pattern types | – | – | √* |
| Output: street infill | Generated street networks | √ | √ | √ |

Note: √ represents that the channel (based on the ground truth) is used in model training. √* denotes the possibility to adopt user-defined inputs.

**Table 2.** Results of mean $l_1$ error and mean $l_2$ error on the entire testing data set.

|  | Mean $l_1$ err. | Mean $l_2$ err. |
|---|---|---|
| Model 1 | 6.55% | 5.94% |
| Model 2[1] | 5.58% | 5.13% |
| Model 2-S1-30%[2] | 5.90% | 5.52% |
| Model 2-S2-60%[2] | 5.56% | 5.09% |
| Model 2-S3-90%[2] | 5.21% | 4.67% |
| Model 3[1] | 5.37% | 4.96% |

Note: [1] represents that the percentage of junction maintained within the generation region for each data sample in both the training and testing stage is randomly generated in range 10% to 90%. [2] represents that three sensitivity tests for Model 2 with 30%, 60%, and 90% of junctions randomly maintained within the generation region for each data sample in both the training and testing stage.

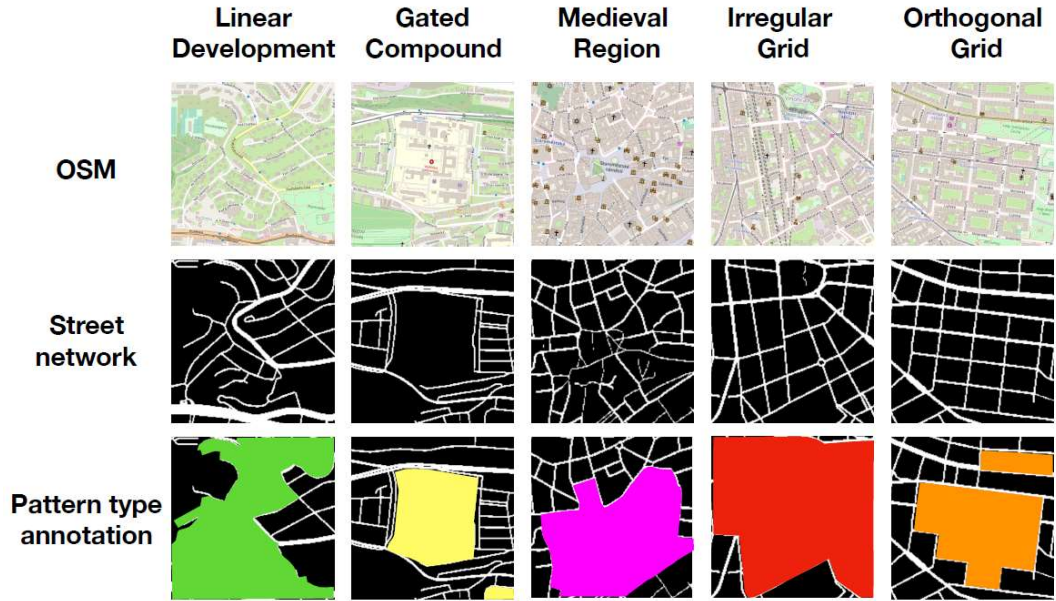

**Figure 1.** Defined street patterns and their representation in Prague.

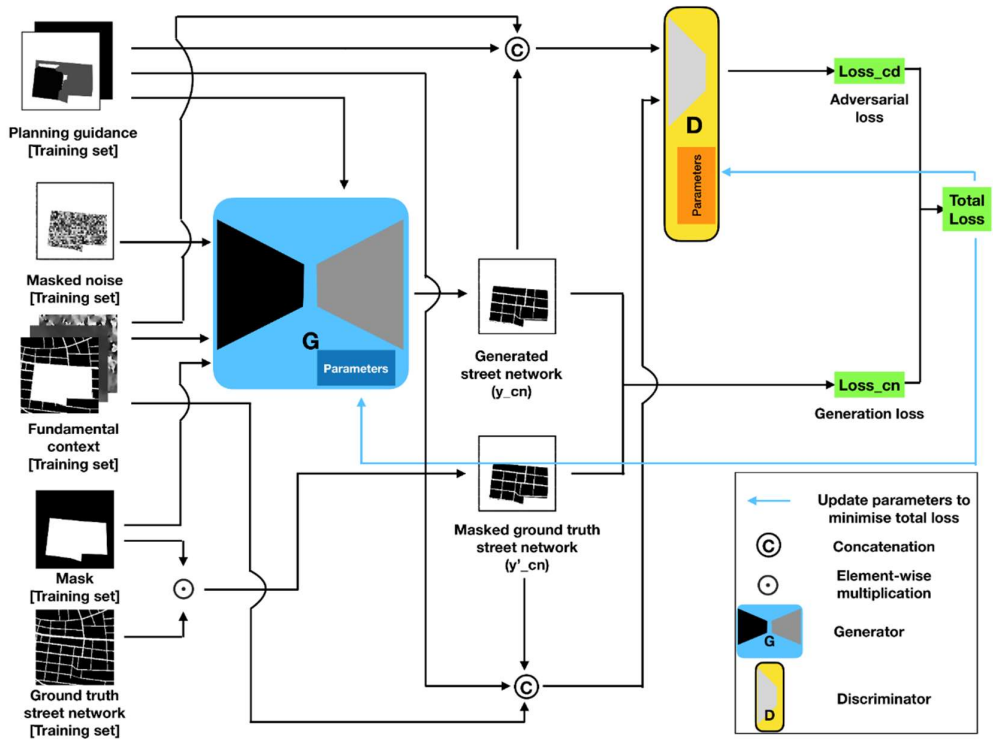

**Figure 2**. Overview of our framework for street network generation.

Note: Detailed network architectures for generator and discriminator are included in the supplementary material.

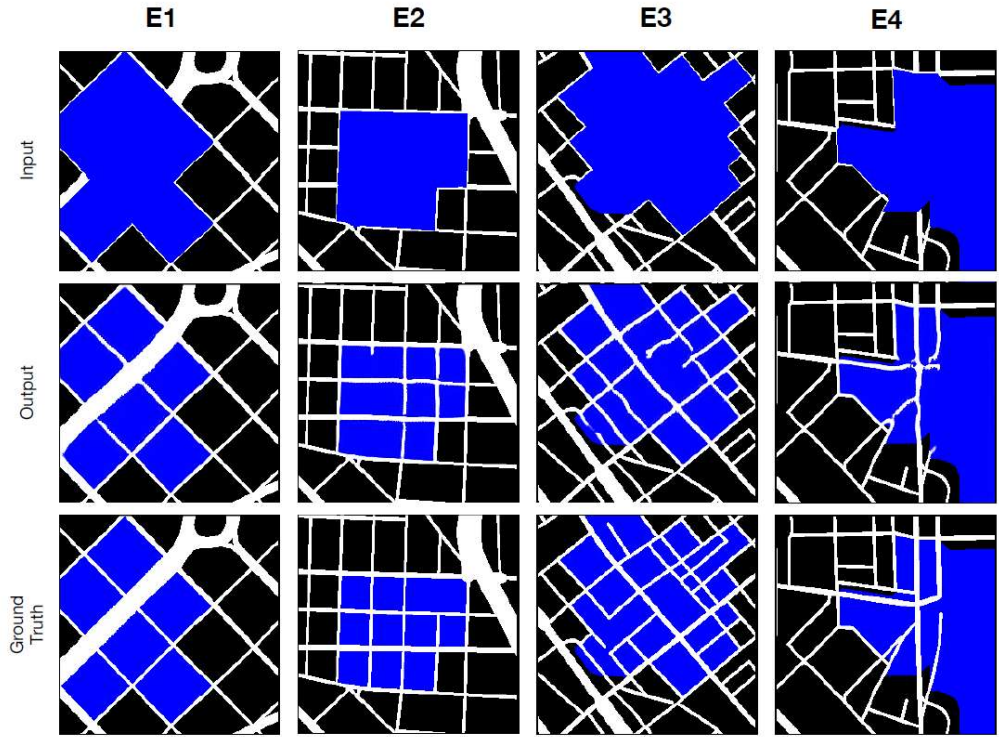

**Figure 3.** Street network generation in Barcelona (Model 1).

Note: Blue area represents the predefined generation region.

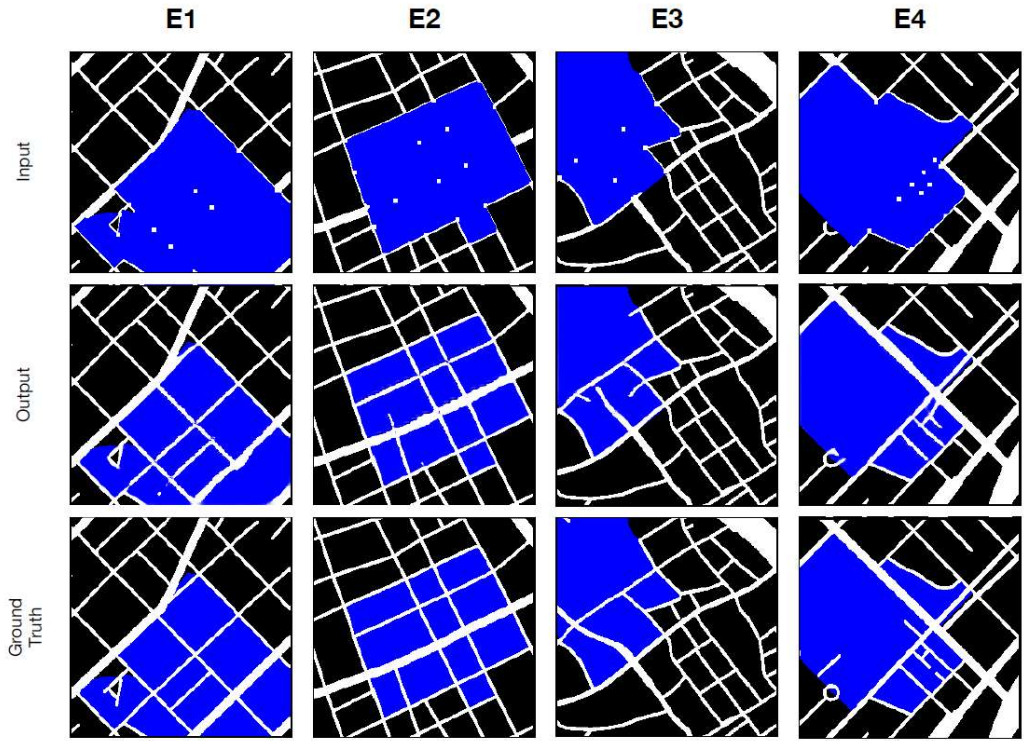

**Figure 4.** Street network generation in Barcelona (Model 2).

Note: Blue area represents the predefined generation region. White dots in the input image represent the randomly maintained junction nodes within the generation region.

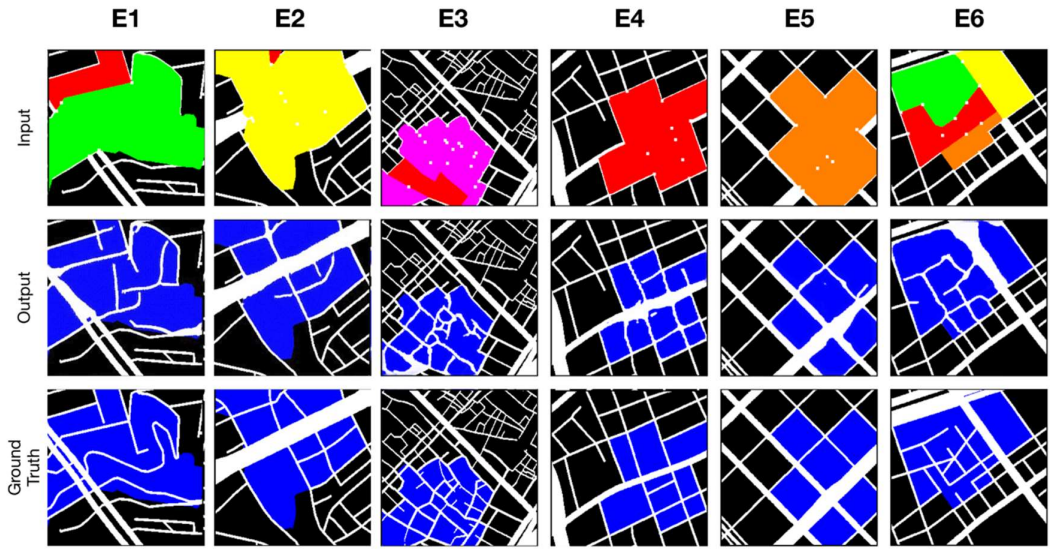

**Figure 5.** Street network generation in Barcelona (Model 3).

Note: Street pattern type: Green–linear development, Yellow–gated compound, Magenta–medieval, Red–irregular grid, Orange–orthogonal grid.

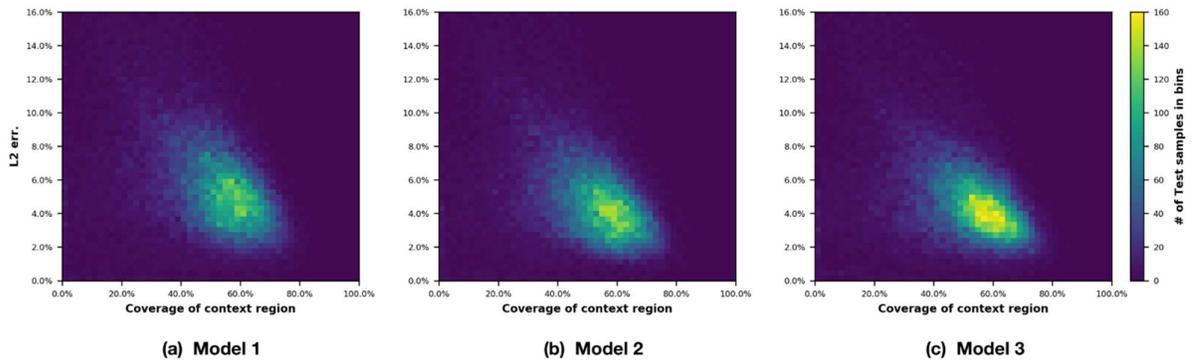

**Figure 6.** Correlations between the coverage of the context region and the results of $l_2$ errors of all samples in the testing set.

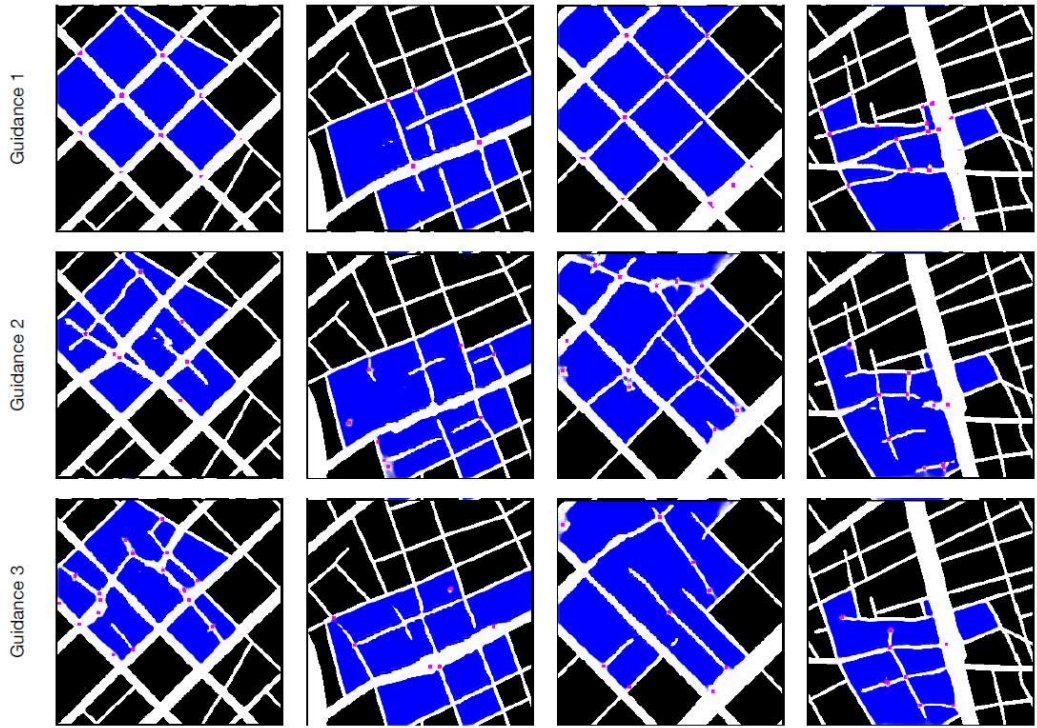

**Figure 7.** The impact of user-defined junction nodes on street network generation.

Note: Blue areas represent the predefined generation regions. Magenta dots denote randomly inserted junction nodes within the generation region.

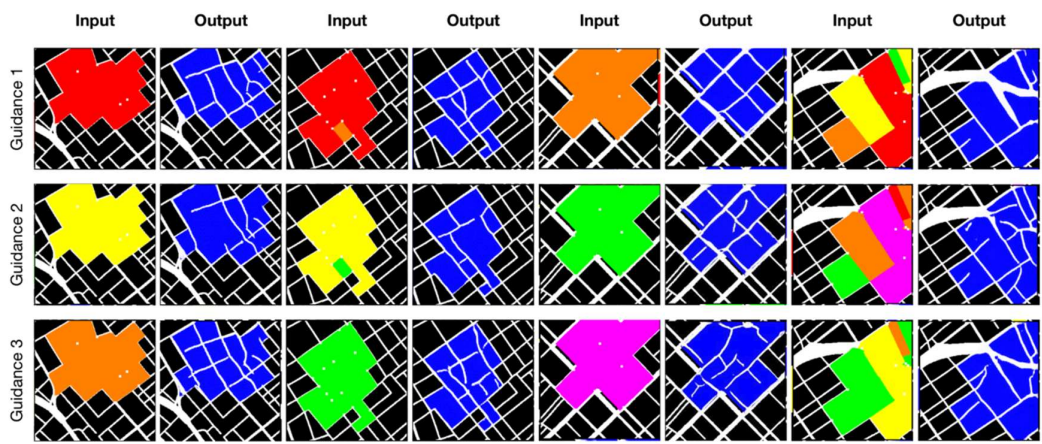

**Figure 8.** The impact of user-defined urban patterns on street network generation.

Note: The color code used here follows the one defined in Figure 5.

**Supplementary materials**

The overall process to set up the database for model training and testing is summarised in Section A. Section B introduces the network architectures of proposed generator and discriminator. This is followed by the details of the adopted loss function and optimization algorithm in Section C. We finally show more street generation results of our models in Section D.

   A. **Database preparation**

      The overall process to set up our database consists of two stages: multi-channel map preparation and data sample preparation, shown in Figure S1 and S2 respectively. In the multi-channel map preparation stage, we first collect multi source data, including open-source vector representation of street network with street level attributes provided by OpenStreetMap (OSM) and public accessible digital elevation data from the Shuttle Radar Topography Mission (SRTM). Then, project these data to the local geo-coordinate system, raster, scale and process them into three single channel raster images: one binary images (street network) and two grey-scale raster images (digital elevation and aspect of slope), where each pixel represents an area of 2×2 meters. Finally, concatenate above single channel images, sharing the same extend, geo-coordinate system and resolution, to a manually labelled map-like street pattern type annotation image to reproduce a Multi-channel map.

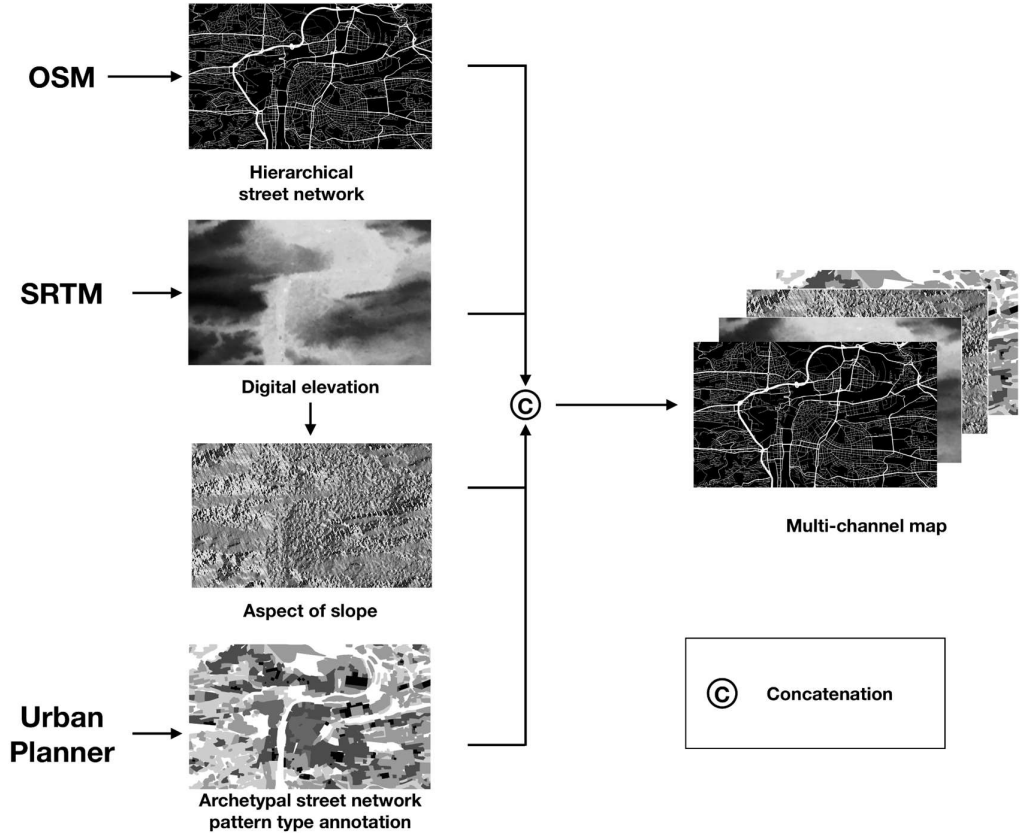

**Figure S1.** Multi-channel map preparation

In the data sample preparation stage, shown in Figure S2, we randomly crop multi-channel patches of size 256×256, from the prepared multi-channel maps and randomly generate binary masks for the cropped patches to define the context region and the region to generate street network. We produce additional same size patch containing two channels for the user guided generation functions of the proposed model for each data sample, one is randomly maintained junction nodes channel and the other is masked pattern type annotation channel. The re-structured fundamental context data sample, planning guidance data samples and mask samples are feed into the proposed neural network pair by pair to train the model.

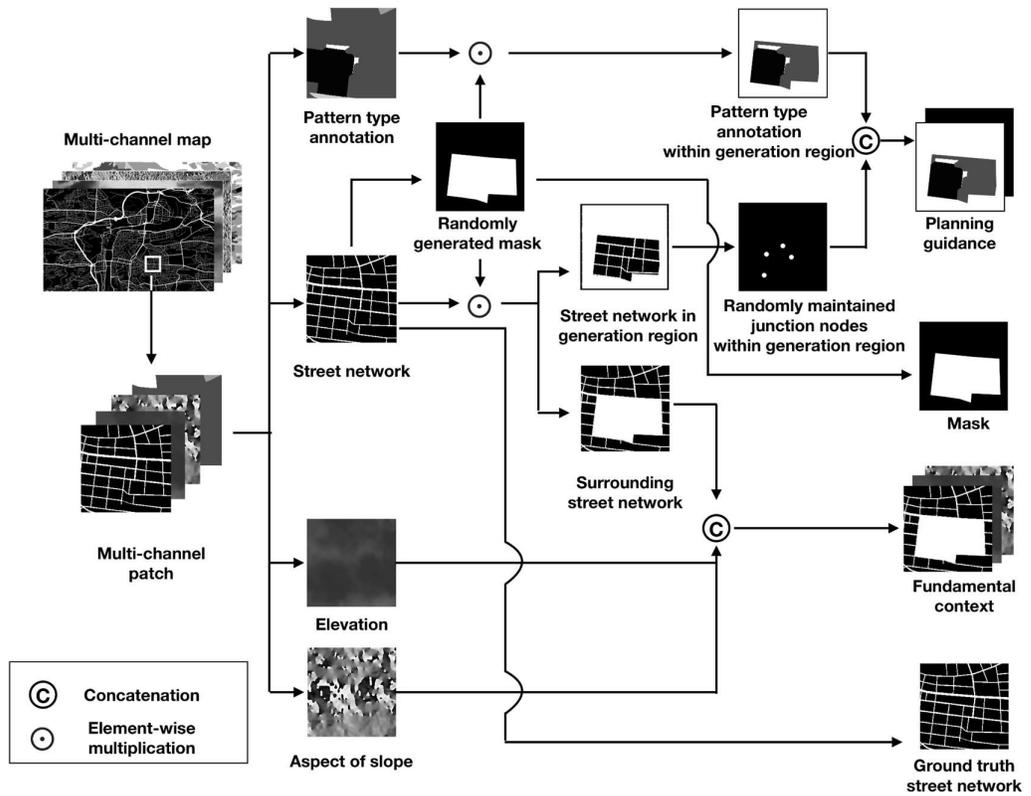

**Figure S2.** Data sample preparation

B. **Network architectures**

The deep neural network proposed for training our model is designed based on the network developed for the image completion task and cGANs. Our network structure is made up of two types of networks: generation network and adversarial network. The generation network, also known as generator, is designed to generate street network within the generation region defined by the mask channel, conditioned on the provided fundamental context and the information stored in the planning guidance. The adversarial networks, also known as discriminator, is trained to guide the training process towards a more realistic street network design. An overview of our network architecture is shown in Figure S3(a) and (b).

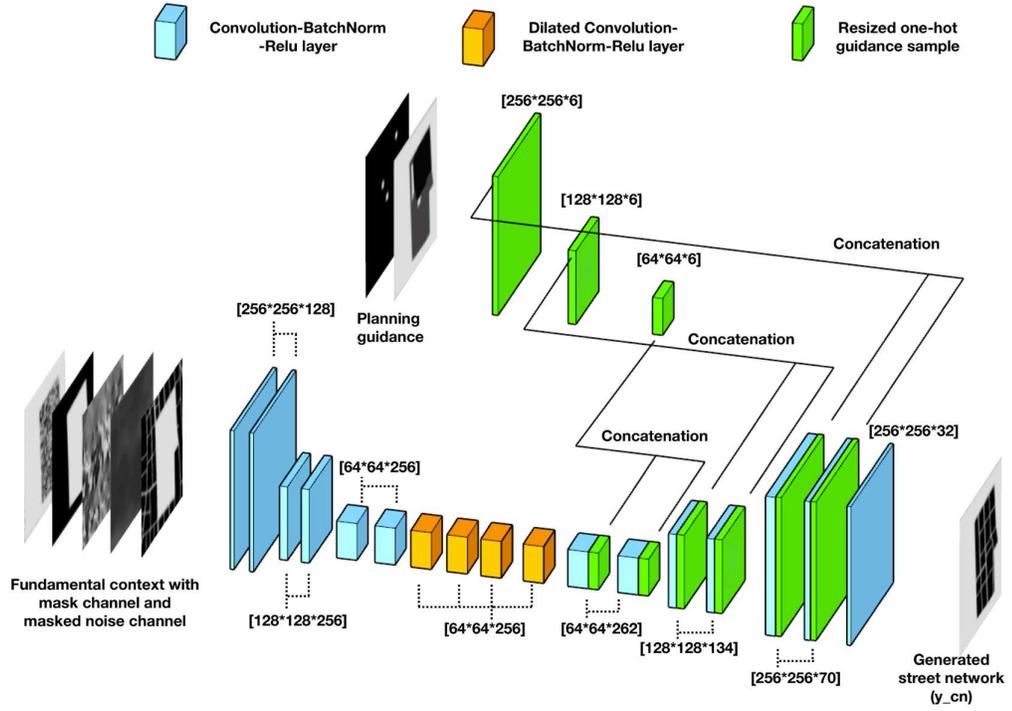

**(a)** Network architecture of generator

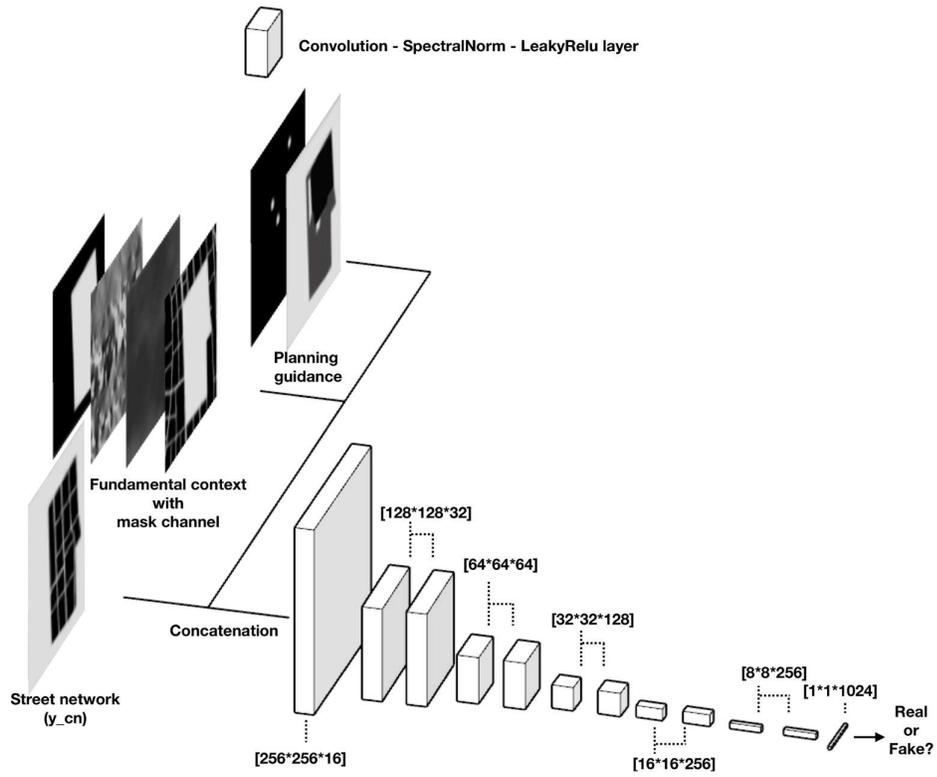

**(b)** Network architecture of discriminator

**Figure S3.** Network architecture of proposed model

*Generation network (Generator):*

The generation network is based on a fully convolutional network. The input of the generation network is a 3-channel fundamental context data sample with a mask channel, masked noise channel and a 2-channel planning guidance data sample. The output is a single channel raster image, represent the generated street network. The general architecture of generator is a classic encoder-decoder structure, which allows for an extracting features in the encoder stage, transforming the context features to predicted features in the fully connected layer, and being restored with repeated concatenated guidance samples to the original resolution in decoder stage using the deconvolution layers (Long et al., 2015). This repetitively concatenations emphasis that the decoding process is conditioned on the information stored in the planning guidance samples.

As the spatial relationships among features are crucial to our street network generation task, the pooling layer, which can reduce image resolution by making the spatial relativity between features less relevant, was not used (Pathak et al., 2016). In our structure, only the stride convolutions are used to decrease the resolution of the input data unit (as shown in Figure 3(a), 256 × 256 to 64 × 64). Through this approach, the spatial relationships among features remain and the computational complexity can be significantly reduced.

Dilated convolutional layers are also used in the mid layers of the generator (highlighted in orange in Figure S3(a)). Dilated convolutions allow for the computing of each output pixel with a much larger input area, while still using the same amount of parameters and computational power (Yu and Koltun, 2015). This is important for our task, as most of the designers would like to design new street network with a deeper understating of the larger extend of surrounding environment.

*Adversarial network (Discriminator):*

A conditional discriminator is attached to the generation network to discern whether a street network is real (ground truth) or generated from generator conditioned on the planning guidance sample. With the conditional discriminators, the generated street network can not only be trained towards realistic texture but also fulfil the pre-requisites, including the forced junction locations and specific street network pattern types. Moreover, as the inputs of the discriminator cover the full extent of samples, including context and generation region of the street network, the generated urban street network can adapt to the street network within the context region and maintain global consistency, and this would mean maintaining quality and character of an urban space, without any threat to urban identity and liveability.

The discriminator takes 1 channel $256 \times 256$ pixels masked street network image concatenated with 3 channels from fundamental context data sample with a mask channel and 2 channels planning guidance data sample as input and transform it through 11 convolution- SpectralNorm -leaky Relu layers and a fully connected layer to an $8 \times 8 \times 256$ vector. This vector is then flattened and processed by another fully connected layer followed by a sigmoid transfer function to a value in the [0,1] range, which represents the probability that the input street network image is real, rather than generated.

## C. Loss function and optimization algorithm adopted

Before introducing the loss function adopted for our model training, the terminology used are listed in Table S1.

**Table S1** Terminology for loss function

| Terminology | Note | Size |
|---|---|---|
| x | Fundamental context data sample with mask channel and masked noise channel | $256 \times 256 \times 5$ |
| Ch_SN | Ch_SN represent the ground truth multi-level Street Network channel for data sample x | $256 \times 256 \times 1$ |
| $x_g$ | Planning guidance sample (masked pattern type annotation channel and masked junction node channel) | $256 \times 256 \times 2$ |
| Mc | Mask channel (1 – within the generation region; 0 – context region) | $256 \times 256 \times 1$ |
| $C(x, x_g)$ | The generation network in a functional form. | The output is a single channel street network image in size $256 \times 256 \times 1$. |
| $D(x, x_g)$ | The adversarial network in a functional form. | The output is in the [0, 1] range and represents the probability that the input single channel street network image is real, rather than generated conditional on $x_g$. |
| $L(x, x_g, M_c, Ch\_SN)$ | $L(x, x_g, M_c, Ch\_SN)$ representing the average pixel wise difference between the generated street network and the ground truth within the generation region calculated follow the Mean Squared Error (MSE) loss, which is defined as $MSE = \frac{1}{n}\sum_{i=1}^{n}(Y_i - \hat{Y}_i)$. | |

In our method, two loss are jointly used to train the module, an MSE loss (used as generation loss), and an adversarial loss. Using the mixture of the two loss functions allows for the stable training of a high-performance network model and has been widely used for image completion.

**MSE loss**: The MSE loss is proposed to quantify the total pixel to pixel difference between the generated street network and the real world one within the generation region. The MSE loss is defined as:

$$L(x, x_g, M_c, Ch\_SN) = \|M_c \odot (C(x, x_g) - Ch\_SN)\|^2$$

where $\odot$ is the pixel-wise multiplication, $\|\cdot\|$ is the Euclidean norm.

**Adversarial loss**: Adversarial loss is considered and included in the loss functions of most image completion algorithms. By taking adversarial loss into account, the standard MSE loss minimization process can be turned into a min-max optimization problem in which the discriminator is jointly updated with the generation network at each training iteration. This is crucial for our approach, due to the existence of multiple possible solutions. In our network, the optimization is:

$$\min_C \max_D E[\log D(M_c \odot x, x_g) + \log(1 - D(C(x, x_g), x_g))]$$

By combining the two loss functions (MSE loss and adversarial loss), the optimization becomes:

$$\min_C \max_D E[L(x, x_g, M_c) + \alpha \log D(M_c \odot x, x_g) + \alpha \log(1 - D(C(x, x_g), x_g))]$$

where α is a weighing hyper-parameter to balance the MSE loss and adversarial loss in the loss function.

For optimization, the ADAM algorithm is used, as this can automatically adapt an appropriate learning rate for each weight in the network (Kingma and Ba, 2015).

## D. Additional results

Finally, we show additional street network generation results for Model 1 to 3 in Figure S4 (a), (b) and (c) and listed additional user guided generation results in Figure S5 (a) and (b).

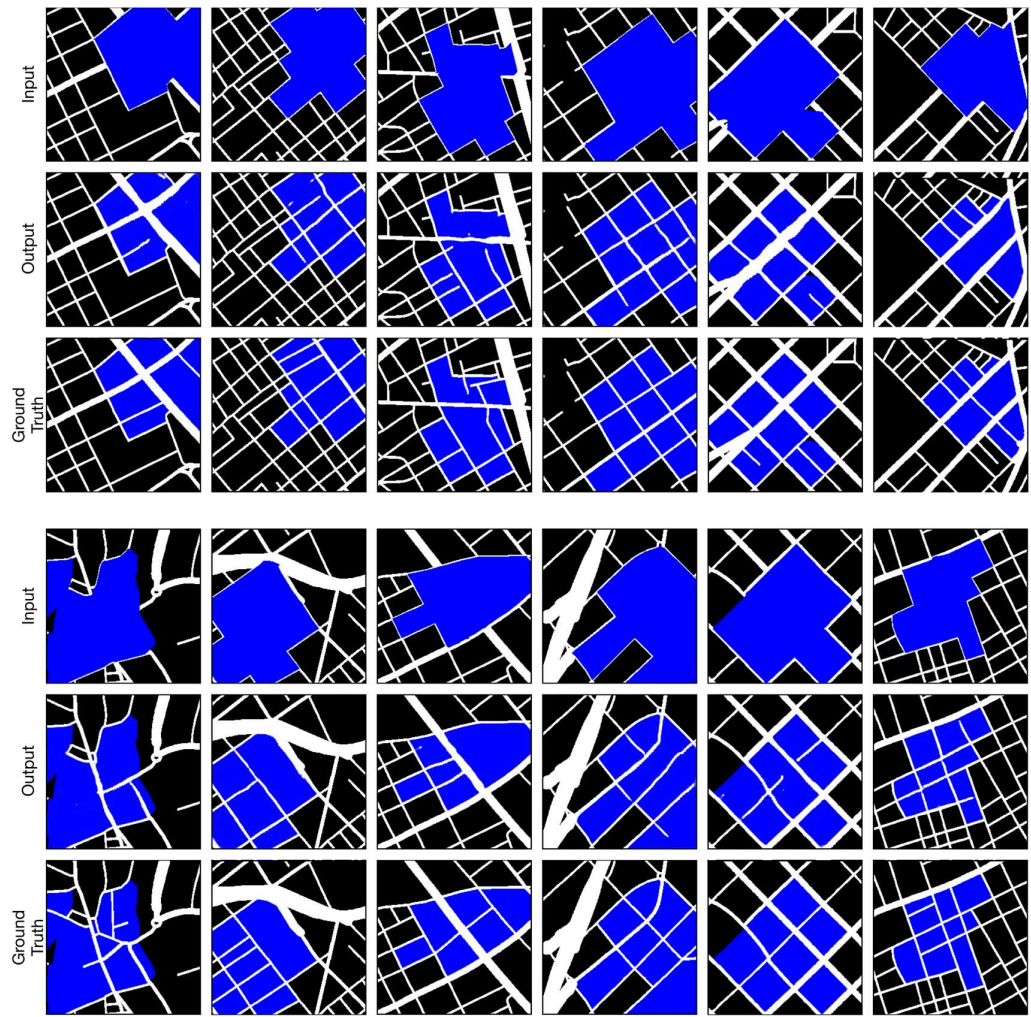

**(a)** Additional generation results of Model 1

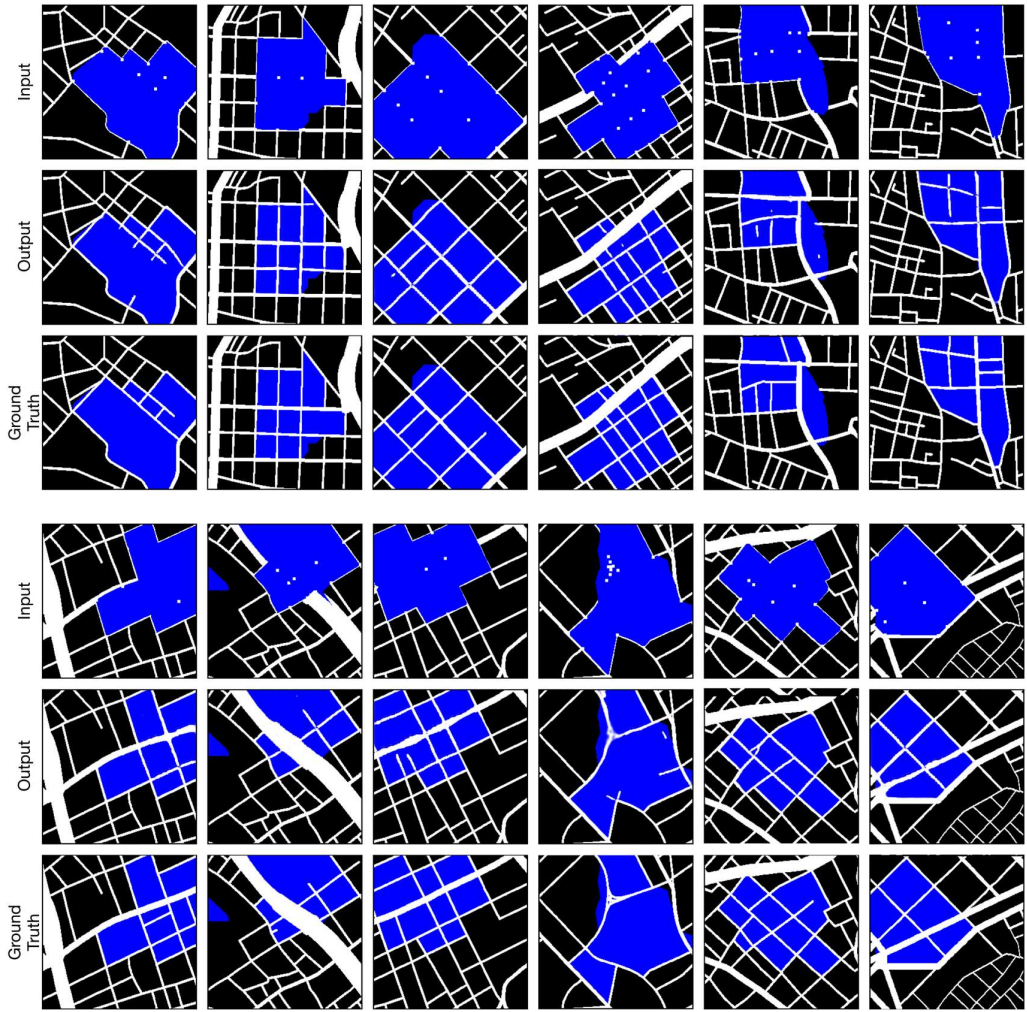

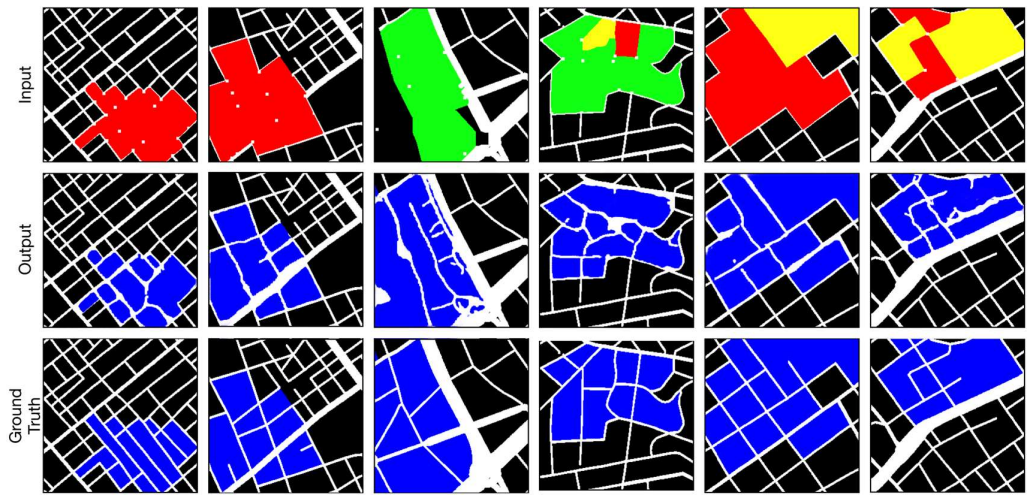

**(b)** Additional generation results of Model 2

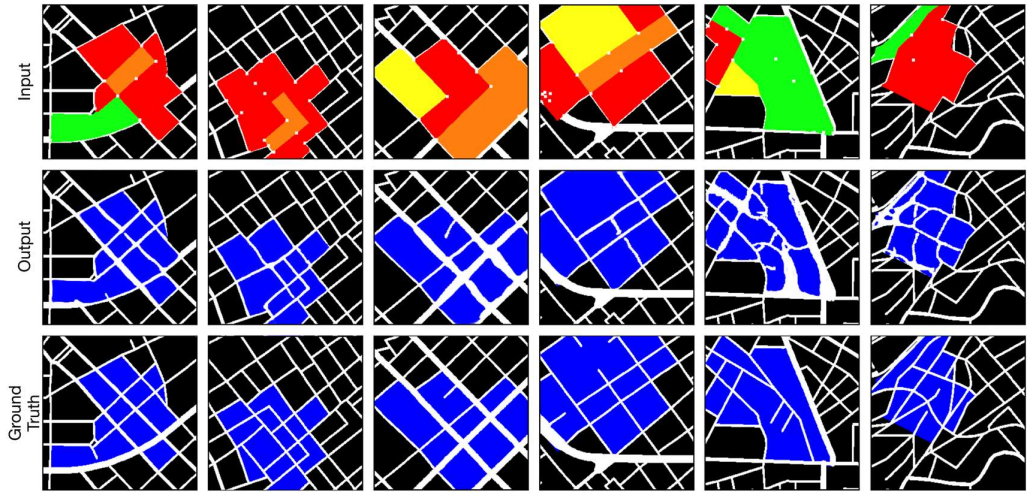

**(c)** Additional generation results of Model 3

**Figure S4.** Additional results for designed experiments (Model 1 to 3)

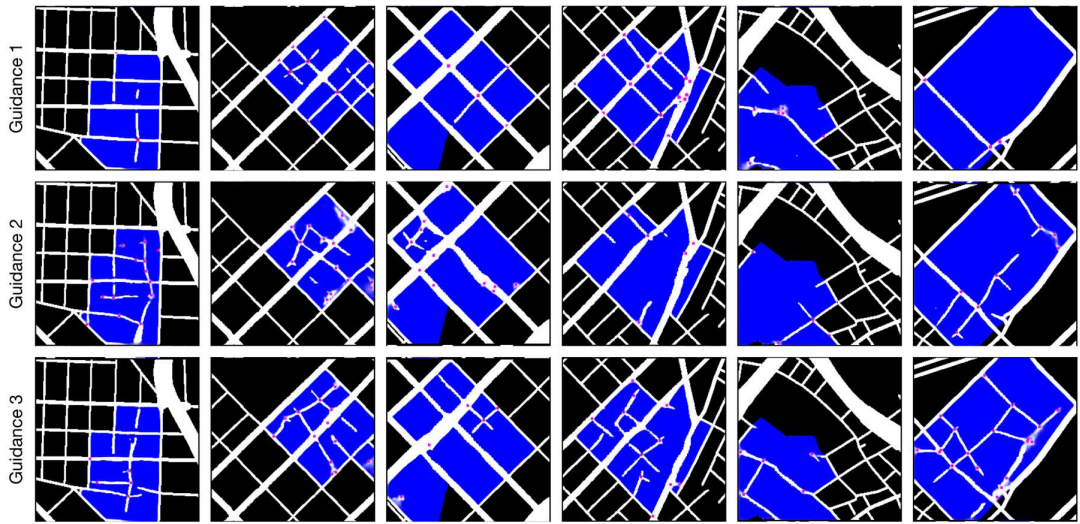

**(a)** Additional user guided generation results (Model 2)

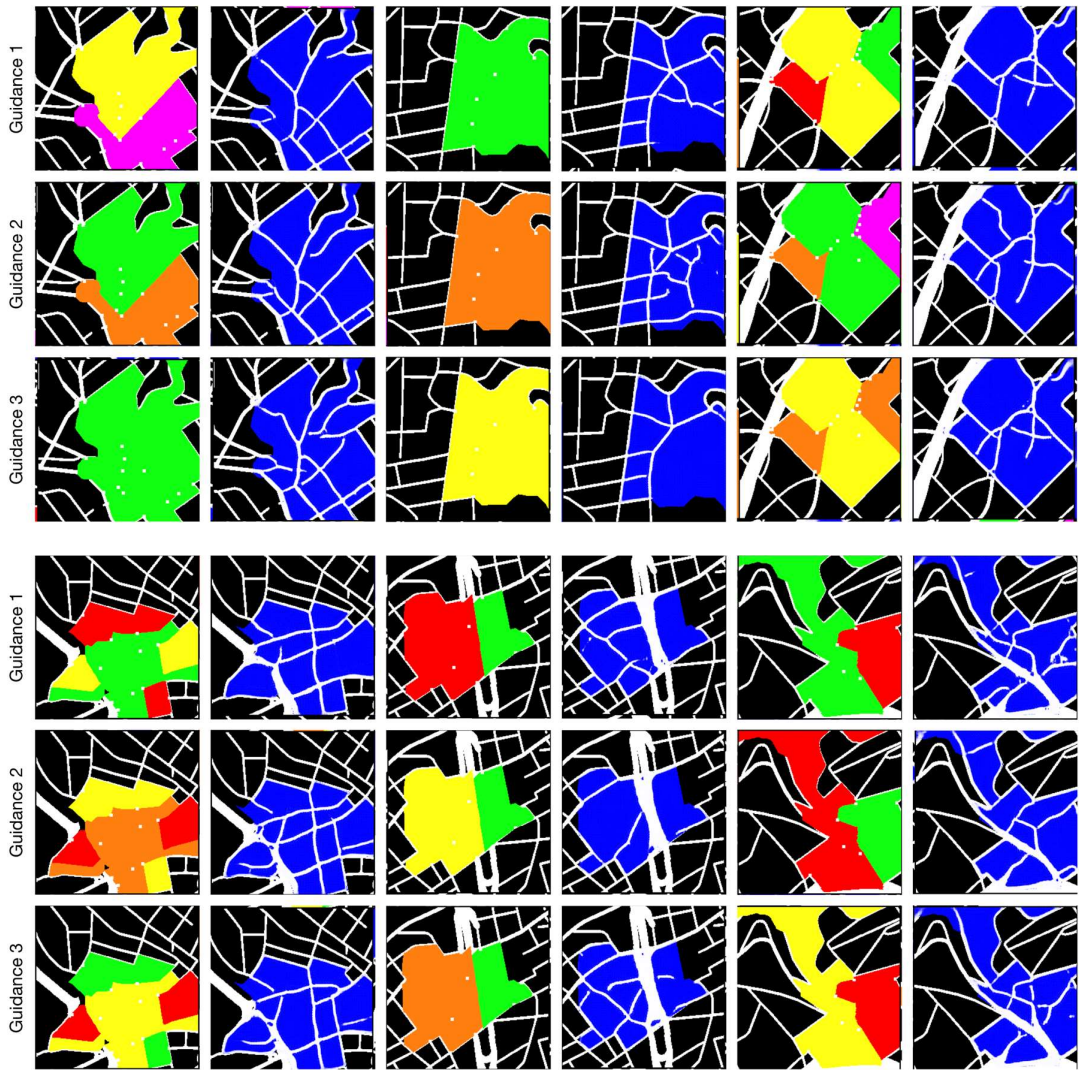

**(b)** Additional user guided generation results (Model 3)

**Figure S5.** Additional results of planning intelligence incorporated generation

**Additional Reference:**